\documentclass[9pt,shortpaper,twoside,web]{ieeecolor}
\usepackage{generic}
\usepackage{cite}
\pdfoutput=1
\usepackage{amsmath,amssymb,amsfonts}
\usepackage{algorithmic}
\usepackage{graphicx}
\usepackage{textcomp}
\usepackage{tabularx}
\usepackage{multirow}
\usepackage{CJK}
\usepackage[justification=centering]{caption}
\def\BibTeX{{\rm B\kern-.05em{\sc i\kern-.025em b}\kern-.08em
    T\kern-.1667em\lower.7ex\hbox{E}\kern-.125emX}}
\begin{document}
\begin{CJK*}{UTF8}{gbsn}
\title{How can Deep Learning Retrieve the Write-Missing Additional Diagnosis from Chinese Electronic Medical Record For DRG}
\author{Shaohui Liu, Xien Liu, Ji Wu
\thanks{Shaohui Liu is an engineer in the electronics department of Tsinghua University(e-mail:shaohuiliu@mail.tsinghua.edu.cn).}
\thanks{Xien Liu is a postdoctoral fellow in the Department of Electronics at Tsinghua University(e-mail:xeliu@mail.tsinghua.edu.cn).}
\thanks{Ji Wu is a professor of electronics at Tsinghua University(e-mail:wuji\_ee@mail.tsinghua.edu.cn).}}

\maketitle

\begin{abstract}
The purpose of write-missing diagnosis detection is to find diseases that have been clearly diagnosed from medical records but are missed in the discharge diagnosis. Unlike the definition of missed diagnosis, the write-missing diagnosis is clearly manifested in the medical record without further reasoning. The write-missing diagnosis is a common problem, often caused by physician negligence. The write-missing diagnosis will result in an incomplete diagnosis of medical records. While under DRG grouping, the write-missing diagnoses will miss important additional diagnoses (CC, MCC), thus affecting the correct rate of DRG enrollment.

Under the circumstance that countries generally start to adopt DRG enrollment and payment, the problem of write-missing diagnosis is a common and serious problem. The current manual-based method is expensive due to the complex content of the full medical record. We think this problem is suitable to be solved as natural language processing. But to the best of our knowledge, no researchers have conducted research on this problem based on natural language processing methods.

We propose a framework for solving the problem of write-missing diagnosis, which mainly includes three modules: disease recall module, disease context logic judgment module, and disease relationship comparison module. Through this framework, we verify that the problem of write-missing diagnosis can be solved well, and the results are interpretable. At the same time, we propose advanced solutions for the disease context logic judgment module and disease relationship comparison module, which have obvious advantages compared with the mainstream methods of the same type of problems. Finally, we verified the value of our proposed framework under DRG medical insurance payment in a tertiary hospital.
\end{abstract}


\section{Introduction}
\label{sec:introduction}

The Diagnosis Related Groups (DRG) is an important tool for measuring the efficiency of medical service quality and medical insurance payment. DRG is essentially a case combination classification scheme. And it is a system that divides patients into several diagnostic groups for management based on factors such as age, disease diagnosis, comorbidities, complications, treatment methods, disease severity, outcome, and resource consumption.
DRG has been used by many countries as a means of medical insurance management and payment. Existing DRG grouping schemes include MS-DRG (US), G-DRG (Germany), AR-DRG (Australia), DPC (Japan), CHS-DRG (China), etc.
Specifically, DRG mainly decides to enter a medical record into a specific DRG group through the principal diagnosis and principal surgery, the presence of complications (CC) or serious complications (MCC), and other individual information. The presence of complications (CC) or severe complications (MCC) was determined by the relationship of additional diagnoses to the principal diagnosis.
The write-missing diagnosis is a very important cause of errors in additional diagnoses, often caused by physician negligence. Unlike the definition of missed diagnosis, a write-missing diagnosis of a disease is clearly manifested in the medical record without further reasoning.
According to some research and actual data, 20\%-30\% of Chinese medical records are missing disease records; in a study of a tertiary hospital in Melbourne, Australia, it was found that missing records of other diagnoses accounted for about 29\% of the wrong medical records in DRG[1].

When countries generally start to adopt DRG enrollment payment, the problem of missing diagnoses is a common and serious problem, which will result in incomplete diagnostic records and inaccurate DRG enrollment. Manual detection of this problem by coders is still the main solution at present. However, due to the complex content of the full medical record, this is costly. We believe that this problem is very suitable to be solved by natural language processing techniques, but to our knowledge, no researchers have conducted research based on natural language processing methods for this problem.

Since the goal of write-missing diagnosis detection is to discover diseases that have been clearly diagnosed in the medical records, in order to be trusted by users, the basis in the medical records should be given to achieve interpretability. So the finalized system will prompt for the location of the missing disease. Unlike disease diagnosis prediction and ICD coding, these only recommend possible diagnoses in probabilistic form.

We defined diagnostic miss-writing as a task that required finding all diseases from the fully electronic medical record that were clearly diagnosed but not recorded in the discharge diagnosis list. The input is the full medical record, and the output is all recommended missed diagnoses. 
Considering the interpretability and traceability issues, we design a framework in the form of a pipeline to solve the problem of diagnosing missing writes.%
\begin{figure*}[!t]
\centerline{\includegraphics[width=13cm]{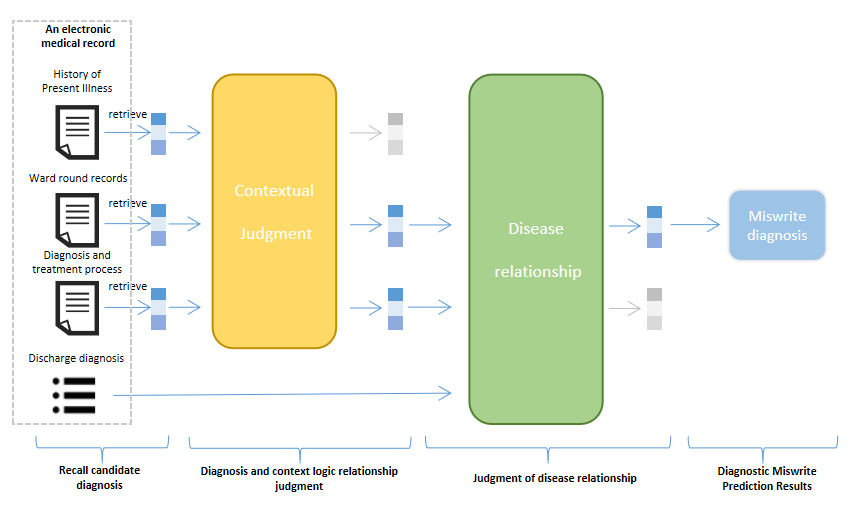}}
\caption{Pipeline framework}
\label{fig1}
\end{figure*}

The pipeline framework we designed mainly includes three sequential modules: the disease recall module, the disease context logic judgment module, and the disease relationship comparison module. The overall architecture is shown in figure \ref{fig1}.
Among them, the disease recall module is mainly responsible for recalling the diseases that appear in medical records, mainly using vocabulary, rules, and named entity recognition methods. 
Because of the complexity of medical record description, not all a disease that appears in the medical record is recorded by the clinician as a diagnosis. Clinicians use the context of the disease to describe the reasons for their records, possibly identifying, preventing, confirming, introducing, etc. Therefore, we designed the disease context logic judgment module to be mainly responsible for judging the logical relationship between the recalled disease and the context, so as to ensure that the candidate missed diseases output in this step are all confirmed diseases this time. The input of this module is all the diseases recalled in the previous step, and the logical relationship between each disease and its context is judged, and finally, all the diseases confirmed this time are selected for output.
After obtaining all the diagnosed diseases this time, it is necessary to compare them with the diseases that have been recorded in the discharge diagnosis list, and select the unrecorded diseases to output as the final write-missing diagnoses. We call this module the disease relationship comparison module, which is responsible for matching two diseases one by one to filter diseases that are similar to the discharge diagnosis list.

Using this framework, we implemented a diagnostic missed write detection system and verified the effect in a tertiary hospital. We took one month's data, compared the use of our diagnostic write-missing detection system, and verified the effect under the local CHS-DRG medical insurance policy. After manual annotation and confirmation, by using our diagnostic missed writing system, 7\% of the cost loss caused by diagnostic missed writing will be avoided.

\subsection{Contributions}

1. Aiming at the problem of write-missing diagnosis detection, a solution framework based on natural language processing technology is proposed for the first time. And the effectiveness of the framework is verified in actual data, and it is also proved that this problem can be effectively solved by natural language processing technology.

2. For each module in the framework, there are significant improvements are designed and compared with the mainstream models, which proves the effectiveness and innovation of our solution to these problems.

3. Apply this write-missing diagnosis detection system to a tertiary hospital that adopts DRG payment, and verify its actual effect through one month's electronic medical record data (EMR). It is found that the use of this system can effectively avoid the 7\% cost loss caused by the diagnosis of missed writing.

\section{Related Work}

\subsection{Related Work for DRG in Medical Information Statistics}
Some researchers related to medical information statistics have analyzed the causes of DRG errors. For example, the researchers verified the influence of the main diagnosis and the omission of other diagnoses on the DRG grouping in a Melbourne tertiary hospital\cite{cheng2009risk}.
\subsection{Related Work for DRG in Natural Language Processing}
Some researchers have used natural language processing technology to study some related tasks involving electronic medical records and DRG. For example, the researchers retrieved the data from 1999 to 2013 from Taiwan's National Health Insurance Research Database (NHIRD) for preliminary diagnosis and prediction of DRG grouping\cite{islam2021deepdrg}.

\subsection{Related Work for Submodule}
Since the technology used in the disease recall module can achieve good results with simple vocabulary matching, we do not focus on it.
ICD automatic coding is a DRG-related task, which has been studied by many scholars. Some researchers used RNN to encode Document and ICD titles respectively and then compared each diagnosis with all ICD titles by attention descriptions \cite{shi2017towards}. Some researchers used Tree-of-sequences LSTM for coding and adversarial learning to enhance model performance \cite{xie2018neural}. Some researchers propose to use hierarchical GRU to encode words and sentences at the same time, through the comprehensive learning of different levels of attention \cite{baumel2018multi}. Interpretability is very important in medical tasks. Some researchers have proposed new methods by capturing the semantic features of discontinuous words and continuous N-gram words, focusing on explaining the reasons for predicting each ICD code \cite{cao2020clinical}.

\textbf{The logical judgment of disease context} can be considered as a natural language inference task. 
Natural Language Inference has been studied for years. Despite lots of work constructing representations for the input of two sentences individually \cite{bowman-etal-2015-large}\cite{mueller2016siamese}\cite{conneau-etal-2017-supervised}, the task actually requires a system to recognize alignments \cite{maccartney2009natural}. In early days, alignment detection is sometimes formed as an independent task\cite{chambers2007learning}\cite{maccartney2008phrase},
or a component of a pipeline system \cite{maccartney2006learning}. Currently, deep learning methods seek to model alignments implicitly through co-attention
mechanism \cite{Parikh2016ADA}\cite{pang2016text}\cite{Chen2016EnhancedLF}\cite{Wang2017BilateralMM}\cite{Gong2017NaturalLI}\cite{Joshi2018pair2vecCW}\cite{DBLP:conf/naacl/DevlinCLT19}. The
technique is first proposed in machine translation \cite{Bahdanau2014NeuralMT}, and soon dominates in many applications including NLI. However why models with co-attention layers are effective is still called for answers.

\textbf{Disease relation comparison} can be considered as a short text-matching task.
Deep Text Matching Models based on deep learning have been widely adopted for short text matching. They can fall into two categories: representation-based methods \cite{he2016text}\cite{lai2019lattice} and interaction-based methods \cite{Wang2017BilateralMM}\cite{Chen2016EnhancedLF}.
Most representation-based methods are based on Siamese architecture, which has two symmetrical networks (e.g. LSTMs and CNNs) to extract high-level features from two sentences. Then, these features are compared to predict text similarity. Interaction-based models incorporate interactions features between all word pairs in two sentences. They generally perform better than representation-based methods.

\section{Task Definition and Dataset}
\subsection{Task Definition}
The formal definition takes the full medical record text $X=[X_1,X_2,...,X_N]$ and the discharge diagnosis $D=[D_1,D_2,...,D_M]$ as the input. Let N is the size of the words of an electric medical record X and M is the size of the discharge diagnosis. The recommended write-missing diagnosis $Y=[Y_1,Y_2,...,Y_T]$ as the output, and let T is the size of recommended write-missing diagnosis. The conditional probability P is described using conditional probability as $P(Y|X,D)$.
If a candidate disease is considered to be missed in the diagnosis, two conditions must be met: one is a confirmed disease existing in the course of the disease, and the other is not recorded in the discharge diagnosis list. This process can be formalized as $P(Y|X,D)=P(Y|Z,D)P(Z|X))$, where $Z$ indicates the confirmed candidate write-missing diagnosis.

Due to the complexity and interpretability requirements of the whole medical record, based on the formal definition, we decompose the recommendation of missed diagnosis into three stages: disease recall, disease context logical judgment, and disease relationship comparison. Among them, disease recall can obtain all diseases from the whole medical record through the disease vocabulary, so we focus on the latter two modules.

\textbf{disease context logic judgment:}The purpose of disease context logic judgment is to judge each input disease $X=[X_1,X_2,...,X_N]$, according to its context $C=[C_1,C_2,...,C_M]$, to judge its logical relationship type Y, that is $P(Y|X,C)$. Let $X_N$ is the size of the input disease in context and M is the size of context. Types of disease-context logic relationships include non-current disease, confirmed disease, and unknown.

\textbf{disease relationship comparison:}The goal of disease relationship comparison is to input two disease names $X_1$ and $X_2$ with N size of characters and determine whether the two are similar, that is $P(Y|X_1,X_2)$. The relationship between diseases includes similarity, inclusion, secondary, and irrelevance.

\subsection{Data Set}
The dataset we use comes from a tertiary hospital in China. The data is the entire medical record, including the medical record home page, admission records, discharge records, disease course records, operation records, death records, and other parts. The data has been desensitized, authorized for scientific research, and subject to confidentiality requirements, so it cannot be disclosed.

Some statistics are as follows in table \ref{Statics}:

\begin{table*}
\caption{Statics of Data Set}
\label{Statics}
\begin{tabular}{|p{70pt}|p{70pt}|p{70pt}|p{70pt}|p{70pt}|p{70pt}|}
\hline
item & number of EMR & number of section & number of diagnosis & average of token length & number of departments\\
\hline
 number & 9405 & 12 & 3126 & 9670.92 & 57 \\
 \hline
 top department & Tumor Chemotherapy Department & Respiratory Medicine & orthopaedics & Internal Medicine-Neurology & Internal Medicine-Oncology\\
\hline
number & 1099 & 761 & 488 & 402 & 373 \\
 \hline
age distribution & newborn & 0-15 & 15-30 & 30-60 & >60\\
\hline
 number & 359 & 1032 & 473 & 4051 & 3490 \\
\hline

\end{tabular}
\label{tab1}
\end{table*}
Since it is very difficult to label the task of missing diagnosis and writing, we invited professional clinicians to label 500 medical records, 400 of which were randomly selected for model training and testing of the disease context judgment and disease relationship comparison modules, and the remaining 100 medical records were used for Final effect evaluation. For disease context judgment, a medical record can generate about 20 pieces of annotated data. For the disease relationship comparison module, a medical record can generate about 60 disease relationship pairs.

The statistical results of the training data set constructed by disease context logic judgment are as follows in table \ref{Statics of Disease Context Logic Judgment}, and the number of test sets is 2,144, and the source is the remaining 100 medical records.
\begin{table}
\caption{Statics of Disease Context Logic Judgment}
\label{Statics of Disease Context Logic Judgment}
\setlength{\tabcolsep}{3pt}
\begin{tabular}{|p{25pt}|p{50pt}|p{75pt}|p{75pt}|}
\hline
item & number of case & average length of context & max length of disease name \\
\hline
 general & 8571 & 654.2 & 2450  \\
 \hline
 labels & non-current disease&confirmed disease& unknown\\
\hline
number & 1363 & 6132 & 1076  \\
\hline
\end{tabular}
\label{tab1}
\end{table}

The statistical results of the training data set constructed by disease relationship comparison are as follows in table \ref{Statics of Disease Relationship Comparison}, and the number of test sets is 1000, and the source is the remaining 100 medical records.
\begin{table}
\caption{Statics of Disease Relationship Comparison}
\label{Statics of Disease Relationship Comparison}
\begin{tabular}{|p{20pt}|p{30pt}|p{30pt}|p{35pt}|p{35pt}|p{35pt}|}
\hline
item & number of case & disease size & average length of disease & min length of disease name & max length of disease name \\
\hline
 number & 25007 &10482 & 12& 1 & 32  \\
 \hline
 labels & similarity& inclusion& secondary&  irrelevance & other\\
\hline
number & 5155 & 8552 & 1371 & 6737 & 3192\\
\hline
\end{tabular}
\label{tab1}
\end{table}

Table \ref{Statics of Write-Missing EMR} shows the overall statistical results for 500 pieces of labeled data that were Write-Missing diagnoses.
\begin{table}
\caption{Statics of Write-Missing EMR}
\label{Statics of Write-Missing EMR}
\begin{tabular}{|p{40pt}|p{100pt}|p{80pt}|}
\hline
 number of EMR & Number of medical records with missed diagnosis  & Number of all missed diseases\\
\hline
500 & 333 & 1213  \\
\hline
\end{tabular}
\label{tab1}
\end{table}

\section{Method}
\subsection{disease context relationship judgment}
After analyzing the data, we found that there are three key problems in the judgment of diagnostic context: 1. diversity of description; 2. Long distance dependence in some cases; 3. There may be multiple descriptions of the same diagnosis in one piece of data, and comprehensive judgment is required. We mainly solve these three problems from the perspectives of data augmentation and models. And I will explain in detail how we solved these problems.

\textbf{data augment}

Because it is difficult to obtain a large amount of annotation data, and the logical relationship between diagnosis and context is very complex, the context representation is extremely rich, which is easy to cause error results in the model in some cases. In one case, because doctors in the same department write in similar ways, the labeled data cannot cover the described situations. Only a few words and words change will make the model difficult to judge accurately. The other case is that the diagnosis in the labeled data has a long tail distribution, and the model is easy to bind the context with the diagnosis in the training process. This makes it difficult for the model to accurately judge the logical relationship between the diagnosis and the context with fewer occurrences.

For this, we use the EDA data enhancement method and the diagnostic name random replacement method to jointly generate enhanced context data. The process is as follows:

\begin{enumerate}
    \item EDA. For a certain piece of data i, the EDA method is used to generate data. In which, the word is taken as the unit to generate twice the data and add it to the training set.
    \item Disease replace. For each piece of data in the training set, randomly replace it with any other disease, and replace all the positions of the disease in the context with new diseases. It should be noted that some diseases, such as diabetes, are difficult to cure completely in medicine, and cannot be replaced with these diseases when they are randomly replaced, so as to ensure that the generated data is still correct. We randomly replaced 3 times disease.
\end{enumerate}

Through data enhancement, we mainly solved the problem of describing diversity.

\textbf{models}
\begin{figure*}[!t]
\centerline{\includegraphics[width=13cm]{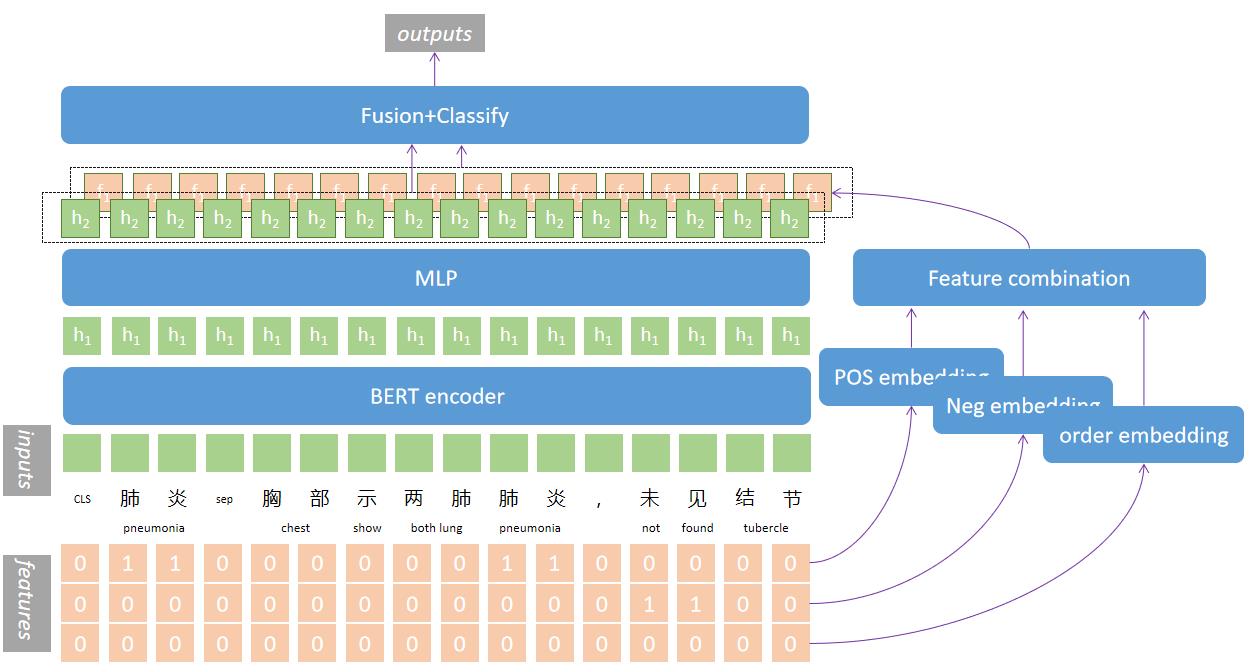}}
\caption{Architecture diagram of disease context logical relationship model}
\label{fig2}
\end{figure*}
Because our annotation data is insufficient, and we need to consider the relationship between diagnosis and the remote context in the judgment process, we choose the BERT model as the research basis, and we can improve the performance by adding features and changing the model structure.

We use the BERT model in the text pair mode. The input is the diagnosis and context. The output is whether the diagnosis and context are positive. In terms of features, we mainly consider several aspects. First of all, since the diagnosis may occur many times in the context, in order to make the model aware of the location of the diagnosis, we use the method of adding features to mark the location where all the diagnosis occurs. Secondly, we collected some different types of words through experience. Their relationship with judgment diagnosis and context logic is very important. We hope that they can increase the judgment ability of the model by taking them as features. Finally, we designed a gating-based method to combine the BERT model coding results with these strong features to predict together, so that the model learning ability has been significantly improved.

The features we use include the following:
\begin{enumerate}
    \item Disease location. For the disease to be judged, retrieve all occurrences in the context. Specifically, set the location where the disease occurs to 1 and other locations to 0.
    \item Negative words. There are many descriptions of negative symptoms in the medical descriptions, which is very important for the logical relationship between disease and context. We first collect the Chinese medical negative word list and then use it in the context. If there are words in the negative word list, they are marked as 1, otherwise, they are 0.
    \item Serial number information. In the actual research process, we found that there are a large number of disease sequences connected by serial numbers in the context, and the diagnosis that needs to be judged is often in such sequences. The relationship between these disease sequences and the context should be consistent, but in fact, the model can not learn this information well, leading to the same disease sequence in the context, where the judgment results of the relationship between each disease and the context are different. In order to make the model learn this information better, we add serial number features to assist. The construction method is to first collect all possible disease sequence forms, then search the disease sequence in the context, and set the position corresponding to the serial number to 1, and the other positions to 0.
\end{enumerate}

The model architecture we use is shown in figure \ref{fig2}. The above strong features are effectively integrated with the output of the BERT model by means of gating, so as to solve the above three problems and improve the performance of the model. The specific calculation method is as follows.

\textbf{Inputs Encoders}

As shown in the figure, $h_1$ represents the result of character input encoded by the BERT model, then $h_1$ is mapped to a specific dimension through a full connection to obtain $h_2$. The calculation method is as follows:
\begin{align}\label{inputs encoder}
    h_1 & = BERT(inputs)\\
    h_2 & = Relu(\mathbf{W_1}^\mathrm{T}h_1+\mathbf{b}_1) 
\end{align}
Where $\mathbf{W_1}$ and $\mathbf{b_1}$ are parameters, $Relu$ is activation function.

\textbf{Features Encoders}

As shown in the figure, the three features are encoded by different embedding lookup tables and then generate $f_1$ through the feature combination layer. The calculation method is as follows:
\begin{align}\label{features encoder}
    &f^{pos} = \mathbf{e}^{pos\quad embedding}(features^{pos})\\
    &f^{neg} = \mathbf{e}^{neg\quad embedding}(features^{neg})\\
    &f^{order} = \mathbf{e}^{order\quad embedding}(features^{order})\\
    &f_1 = Relu(\mathbf{W}_{pos}^{\mathrm{T}}f^{pos}+\mathbf{W}_{neg}^{\mathrm{T}}f^{neg}+\mathbf{W}_{order}^{\mathrm{T}}f^{order}+\mathbf{b}_{f})
\end{align}
Where $features^{pos}$, $features^{neg}$ and $features^{order}$ are respectively the feature inputs of disease location, negative words and serial number information. And $\mathbf{e}^{pos\quad embedding}$, $\mathbf{e}^{neg\quad embedding}$ and $\mathbf{e}^{order\quad embedding}$ are respectively the embedding lookup tables. $\mathbf{W}_{pos}$, $\mathbf{W}_{neg}$ and $\mathbf{W}_{order}$ are respectively the parameters.

\textbf{Fusion \& Classify}

By using the gating mechanism, we use the feature information to effectively enhance and utilize the input context information. Finally, the logical relationship between disease and context is classified by using the classification layer. The calculation method is as follows:
\begin{align}\label{inputs encoder}
    & h_3 = tanh(\mathbf{W}_{fm}^\mathrm{T}[f1;h2]+\mathbf{b}_{fm})\\
    & g = sigmoid(\mathbf{W}_{g}^\mathrm{T}[f1;h2]+c)\\
    & o = g \odot h_3 + (1-g) \odot h_2\\
    & c = [max-pooling(o);mean-pooling(o)]\\
    & y = softmax(\mathbf{W}_{y}^\mathrm{T}c+\mathbf{b}_y)
\end{align}
Where $\mathbf{W}_{fm}$, $\mathbf{W}$, $\mathbf{W}$, $\mathbf{b}_{fm}$, $\mathbf{b}_{fm}$ and $\mathbf{b}_y$ are parameters. The g is the gate generated by inputs information and features information to control the outputs.

\subsection{disease relationship comparison}
For the comparison of disease relationships, the biggest difficulty lies in the huge number of diseases. In ICD-10 (International Classification of Diseases), there are more than 33000 disease codes. The number of diagnoses that may be compared in diagnostic miss write can reach 900,000,000. Compared with such a huge disease scale, the number of doctors' labels is too small, which makes it difficult for the model to fully learn the relationship between diseases. To solve this problem, we designed a special data enhancement method, so that the model can fully learn the relationship between diseases as possible.

In order to enable the model to fully learn the relationship between diseases, it would be inefficient to construct 900000000 disease pairs as training data by direct random sampling, and it would be extremely difficult for the model to learn such a data set. 

Therefore, we design a special pre training data construction method to construct positive and negative case data by using the characteristics of electronic medical record diagnosis writing.

\textbf{Positive example construction}:
\begin{itemize}
    \item Based on the characteristics of medical record writing. The clinical diagnosis name on the first page of the medical record needs to be coded by the coder into the ICD code, which is a process of terminology standardization. Of course, occasionally based on medical records, the coders will code the clinical diagnosis name to a more fine-grained ICD code. Since it is based on human coding, the error rate is often less than 10\%. Therefore, we believe that the relationship between the clinical diagnosis name and the corresponding ICD standard disease name is very similar, which is a similar or inclusive relationship and is used as a positive example in the pre-training task.
    \item Based on common data augmentation methods. EDA (Easy Data Augmentation) and back-translation methods are commonly used data augmentation methods in natural language processing. However, due to the dense information of disease names, the EDA method will generate a lot of noisy data, so we only use the back-translation method to construct positive examples.

\end{itemize}

\textbf{Negative example construction}:
\begin{itemize}
    \item Based on the characteristics of medical record writing. On the first page of the medical record, according to the general requirements for diagnosis writing, clinicians will not fill in two similar diagnoses into one diagnosis list at the same time. Based on this idea, we treat different diagnoses in the same diagnosis list as negative examples of each other. Including clinical diagnosis name and clinical diagnosis, clinical diagnosis name and ICD coding standard disease name, ICD coding standard disease name, and ICD coding standard disease name.
    \item Tree structure based on ICD encoding. The structure of the ICD code is tree-like, such as the 4-digit code S05.3 (ocular laceration without prolapse or loss of intraocular tissue), which has an inclusive relationship with S05.301 (scleral rupture). While S05.3 and S05.4 (penetrating orbital injury with or without foreign body) have the same 3-digit code, the relationship between the two is similar but different. According to this characteristic, we treat diseases with the same 3-digit code, different 4-digit code, or 6-digit code as negative cases of each other.
    \item Based on random sampling. Randomly sample other codes from the ICD codes, generating random negatives.
\end{itemize}


Based on the above methods, we constructed a series of augmented data for disease relationship comparisons from unlabeled medical record data. Because there are some differences between these enhanced data construction methods, doctors' annotation principles and labels number, it is not suitable to directly add training data. In order to make better use of these large amounts of data, we use them as pre-training data so that the model can learn important information. The statistical results are shown in Table \ref{Statics of pretrained data for disease relationship comparisons}.
\begin{table}
\caption{Statics of pretrained data for disease relationship comparisons}
\label{Statics of pretrained data for disease relationship comparisons}
\centering
\begin{tabular}{|c|c|c|}
\hline
 labels & same diseases pair & dissimilar diseases pair\\
\hline
number & 302000 & 410000 \\
\hline
\end{tabular}
\label{tab1}
\end{table}

Therefore, in the phase of disease comparison, the method we use is to build positive and negative data based on medical knowledge and characteristics of medical electronic medical records. Then we use these data for pre-training and finally fine-tune the labeled data. In terms of the model, we use the classical BERT model to make use of the rich knowledge in the BERT model to make the disease relationship comparison model easier to learn. And We use the SimCSE model as a comparative learning method for pre-training.

\section{Experiment}
In our proposed framework, the task of diagnosis missed writing detection mainly studies the two sub-modules of disease context relationship judgment and disease relationship comparison. For the convenience of description, we will describe these two modules and the overall experimental results respectively.

\subsection{disease context relationship judgment}

\textbf{Experiment Settings}

The disease context judgment task is similar to the natural language inference task. We evaluate the performance of our proposed model on the test set and compare it with other classic models for natural language inference tasks by using the labeled data as training and test sets. We use the accuracy, precision, recall, and F1 as the evaluation index and set the BERT-based model for sentence-pair relationship judgment as the baseline model.

\textbf{Hyperparameter Settings}

Because models such as BERT have input length limitations and considering that too long context has little impact on the resulting judgment, we set the maximum length of the context to 450 and the maximum length of the diagnosis to 30. In order to avoid different results caused by different hyperparameter settings, for some key parameters such as Table \ref{Hyperparameter Settings of Disease Context Relationship Judgment}, we set all models the same. For other parameters, it is set according to the paper or public code. To ensure fairness, for each experiment, We experiment three times and choose the best score as the result.
\begin{table}
\caption{Hyperparameter Settings of Disease Context Relationship Judgment}
\label{Hyperparameter Settings of Disease Context Relationship Judgment}
\centering
\begin{tabular}{|c|c|}
\hline
 item & value\\
\hline
batch size & 64 \\
\hline
learning rete & 5e-5 \\
\hline
pre-trained model & BERT-base \\
\hline
max length & 450 \\
\hline
loss & focal loss \\
\hline
\end{tabular}
\label{tab1}
\end{table}
For focal loss, we use the officially recommended parameter, gamma=2.

\textbf{Experiment Result}

\begin{table}
\caption{Experiment Result of Disease Context Relationship Judgment}
\label{Experiment Result of Disease Context Relationship Judgment}
\centering
\begin{tabular}{p{60pt}|c|c|c|c}
\hline

 model & acc & macro-p & macro-r & macro-f1\\
\hline
LSTM &  0.811 & 0.778 & 0.741 & 0.760 \\
BIMPM &  0.832 & 0.818 & 0.742 & 0.778 \\

Bert-base CLS & 0.909 & 0.870 & 0.861 & 0.865 \\

Bert-base Mean \& Max& 0.908 & 0.874 & 0.860 & 0.867\\

\hline
our model & 0.914 & 0.888 & \textbf{0.867} & 0.875 \\

our model + data augment & \textbf{0.921} & \textbf{0.916} & 0.856 & \textbf{0.876} \\
\hline
\end{tabular}
\label{tab1}
\end{table}
As shown in Table \ref{Experiment Result of Disease Context Relationship Judgment}, our model outperforms the baseline by a large margin and significantly outperforms the mainstream models in natural language inference tasks, achieving the best results. Thus the effectiveness of our model is demonstrated. In the table, the LSTM denotes the simple model constructed by LSTM and MLP to classify.

\textbf{Ablation experiment}

\begin{table}
\caption{Experiment Result of Disease Context Relationship Judgment}
\label{Experiment Result of Disease Context Relationship Judgment}
\centering
\begin{tabular}{p{60pt}|c|c|c|c}
\hline
 item & acc & macro-p & macro-r & macro-f1\\
\hline
our model& \textbf{0.921} & \textbf{0.916} & 0.856 & \textbf{0.876} \\
\hline
-EDA & 0.918 & 0.883 & 0.867 & 0.875\\

-Disease replace & 0.914 & 0.888 & 0.867 & 0.876 \\
\hline
-Negative words & 0.914 & 0.880 & \textbf{0.870} & 0.875\\

-Serial number information & 0.912 & 0.875 & \textbf{0.870} & 0.872 \\

-Disease location & 0.908 & 0.874 & 0.860 & 0.867 \\

\hline

\end{tabular}
\label{tab1}
\end{table}
As shown in Table \ref{Experiment Result of Disease Context Relationship Judgment}, the data augment methods and three key features in our model all lead to a decrease in performance after removal, especially feature of Disease location, which has the most significant effect on the performance of the model.

\textbf{Case Study}
\begin{table*}
\caption{Case Study of Disease Context Relationship Judgment }
\label{Case Study of Disease Context Relationship Judgment}
\centering
\begin{tabular}{|p{40pt}|p{250pt}|p{40pt}|p{30pt}|p{30pt}|p{40pt}|}
\hline

 disease & context & relation & LSTM & Bert-base & our models with data augment\\
\hline

肺心病 & 患者诉心慌、胸痛，憋喘，不能平卧，咳嗽无咳痰，有发热症状，大小便正常。就诊于地方医院，考虑考虑肺部感染可能性大，不能除外\textbf{\emph{肺心病}}，予以抗感染治疗后效果欠佳，遂就诊于我院，查心超：左心增大，心包积液，左室收缩功能减低，中度肺动脉高压伴三尖瓣少量返流，二尖瓣大量返流，主动脉瓣中量返流；根据患者病史及相关辅助检查，心电图、X线胸片、超声心动图有右心增大肥厚的征象，现确诊为\textbf{\emph{肺心病}}，心功能不全，肺部感染；治疗上主要予以改善心功能、抗感染、改善呼吸功能等对症治疗，继续观察。 
&
confirmed disease 
& 0.566 & 0.723 & \textbf{0.847}\\
\hline
pulmonary heart disease& The patient complained of palpitation, chest pain, dyspnea, inability to lie flat, cough without expectoration, fever, and normal urine and stool. Then the patient went to a local hospital. Considering that the possibility of pulmonary infection was high, the doctor could not exclude \textbf{\emph{pulmonary heart disease}}, and the effect of anti-infection treatment was not good, so the patient went to our hospital. A cardiac ultrasound examination showed that the left heart was enlarged, pericardial effusion, left ventricular systolic function was reduced, moderate pulmonary hypertension with a small amount of tricuspid regurgitation, a large amount of mitral regurgitation, and a moderate amount of aortic regurgitation; According to the patient's medical history and relevant auxiliary examinations, ECG, X-ray chest film and echocardiography showed signs of enlargement and hypertrophy of the right heart, and now it is diagnosed as \textbf{\emph{pulmonary heart disease}}, cardiac insufficiency and pulmonary infection; In terms of treatment, symptomatic treatment such as improving heart function, anti-infection and improving respiratory function was mainly given, and continued observation was made. & confirmed disease & 0.566 & 0.723 & \textbf{0.847}\\
\hline
\end{tabular}
\label{tab1}
\end{table*}

Table \ref{Case Study of Disease Context Relationship Judgment} shows a common case of patient visits. For the diagnosis of pulmonary heart disease that needs to be judged, there are two descriptions in this example, but the meaning of its local context is very different. The first department said that it had seen a doctor in other hospitals but could not make a diagnosis. The relationship should be "unknown". The second place said that the diagnosis of pulmonary heart disease was finally made after the examination, and the relationship should be "confirmed disease". This requires that the model has the ability to handle complex logic. The model should be able to understand the real logical relationship between diagnosis and its context diagnosis and make comprehensive decisions. In this case, all models tend to predict the confirmed disease, and our model has the highest confidence in prediction. Our guess is that the diagnostic location feature enables the model to directly locate all the locations where the diagnosis occurs, and it is easy to learn the sequence of multiple expressions in the sentence.

\subsection{disease relationship comparison}
\textbf{Experiment Sets}

The disease relationship comparison task can be considered as a text matching task, so we select some excellent models in the field of text matching for comparison with our model. We choose the accuracy, precision, recall and f1 as the evaluation metric, and use the BERT-based sentence pair matching model as the baseline.

\textbf{Hyperparameter Settings}

According to the statistical results, we set the longest input of disease to 50. In order to avoid different results caused by different hyperparameter settings, for some key parameters such as Table \ref{Hyperparameter Settings of disease relationship comparison}, we set all models the same. For other parameters, it is set according to the paper or public code. To ensure fairness, for each experiment, We experiment three times and choose the best score as the result.
\begin{table}
\caption{Hyperparameter Settings of Disease Relationship Comparison}
\label{Hyperparameter Settings of disease relationship comparison}
\centering
\begin{tabular}{|c|c|}
\hline
 item & value\\ 
\hline
batch size & 256 \\
\hline
learning rete & 5e-5 \\
\hline
pretrained model & BERT-base \\
\hline
max length & 50 \\
\hline
loss & cross entropy\\
\hline
$\tau$ for simCSE  & 0.05\\
\hline
\end{tabular}
\label{tab1}
\end{table}
It should be noted that when using simCSE for comparative training, in order to avoid catastrophic forgetting of important information in BERT, we set the learning rate to 1e-6.

\textbf{Experiment Result}

\begin{table}
\caption{Experiment Result of disease relationship comparison}
\label{Experiment Result of disease relationship comparison}
\centering
\begin{tabular}{p{60pt}|c|c|c|c}
\hline
model & acc & macro-p & macro-r & macro-f1\\
\hline
LSTM & 0.685 & 0.651 & 0.638 & 0.641 \\

BIMPM & 0.721 & 0.713 & 0.678 & 0.695 \\

Bert-base-CLS & 0.821 & 0.770 & 0.762 & 0.764 \\

\hline
Bert-base-CLS with pretrain & \textbf{0.853} &  \textbf{0.831} & \textbf{0.785} & \textbf{0.804} \\
\hline
\end{tabular}
\label{tab1}
\end{table}
As shown in Table \ref{Experiment Result of disease relationship comparison}, Bert-base-CLS with pretrain that we propose outperforms the baseline has significant improvement the mainstream models in text-matching tasks, achieving the best results. Thus the effectiveness of our model is demonstrated. 

\textbf{Ablation experiment}
\begin{table}
\caption{Experiment Result of disease relationship comparison}
\label{Experiment Result of disease relationship comparison}
\centering
\begin{tabular}{p{60pt}|c|c|c|c}
\hline
Bert-base-CLS with pretrain & \textbf{0.853} &  \textbf{0.831} & \textbf{0.785} & \textbf{0.804} \\
\hline
-common data augmentation methods & 0.849 & 0.817 & \textbf{0.785} & 0.800\\

-Tree structure based on ICD encoding & 0.835 & 0.801 & 0.771 & 0.786\\

-characteristics of medical record writing & 0.821 & 0.770 & 0.762 & 0.764 \\
\hline

\end{tabular}
\label{tab1}
\end{table}
As shown in Table \ref{Experiment Result of disease relationship comparison}, our data enhancement method effectively improves the effect of the pre-training contrast model. Among them, the method based on characteristics of medical record writing constructs positive and negative cases based on the characteristics of the medical record itself, which is helpful for the results.

\textbf{Case Study}

\begin{table*}
\caption{case of disease relationship comparison}
\label{case of disease relationship comparison}
\centering
\begin{tabular}{|p{60pt}|p{60pt}|c|c|c|c|c|}
\hline
 disease1 & disease2 &relation& Edit distance & LSTM & Bert-base & ours\\
\hline
头部骨折(Head fracture) & 头骨骨折(skull fracture) & similarity & similarity(0.75) & \textbf{similarity(0.985)}& similarity(0.945) & similarity(0.923)\\
\hline
肺炎(pneumonia) & 肺部感染(pulmonary infection) & similarity & similarity(0.333) & similarity(0.553) & similarity(0.776) & \textbf{similarity(0.911)}\\
\hline
电解质紊乱(Electrolyte disturbance) & 低钾血症(Hypokalemia) & inclusion & inclusion(0) & inclusion(0.003) & inclusion(0.223) & \textbf{inclusion(0.635)}\\
\hline
\end{tabular}
\label{tab1}
\end{table*}

Some examples are shown in the table \ref{case of disease relationship comparison}. The scores in parentheses represent the scores of the correct type predicted by the model. It can be seen from the table that for simple types, such as the relationship between head fracture and skull fracture, all methods can be used to correctly predict. For disease pairs that require shallow semantic knowledge to judge, such as pneumonia and lung infection, most models can correctly predict through training. The BERT model can predict such problems very well because it has a lot of knowledge obtained from pre-training data. For diseases such as electrolyte disorder and hypokalemia, because the training set is too small compared with the whole set of diseases, it is difficult to obtain ordinary models through training and one must have more professional knowledge. Our model can better solve such problems by learning the real disease code from medical record data and ICD-10 knowledge.

\subsection{overall}
\textbf{Experiment Sets}

Based on the framework proposed in the model chapter, we connected the vocabulary-based disease name recall module, the diagnosis context relationship judgment module, and the disease relationship comparison module in series according to the pipeline form to realize the diagnosis missed writing system. To this end, we have collected a disease vocabulary of about 40000 for recalling diseases. Only when the judgment result of the disease context is confirmed, the judgment of the disease relation module will be performed. Finally, when a candidate diagnosis is judged as irrelevant by the disease relationship comparison module with all discharged diagnoses, it will be output as a missed writing diagnosis. We also analyzed the value generated by the diagnostic context judgment module and the disease comparison module from an overall perspective. We use precision, recall, and F1 value to measure the performance of the missed diagnosis detection system.

\textbf{Experiment Result}

\begin{table}
\caption{Experiment Result of overall}
\label{Experiment Result of overall}
\centering
\begin{tabular}{p{100pt}|p{30pt}|p{20pt}|p{20pt}}
\hline
 item & precision &recall& F1\\
\hline

overall &  \textbf{0.925}&0.715&\textbf{0.807} \\
\hline
overall without diagnosis context relationship judgment module & 0.652&\textbf{0.823}&0.728 \\
\hline
overall without relationship comparison module & 0.692&0.799&0.742 \\
\hline
\end{tabular}
\label{tab1}
\end{table}
It can be seen from Table \ref{Experiment Result of overall} that the diagnostic context judgment and disease relationship comparison modules are both necessary modules, which are indispensable.

\subsection{Reality Effect Verification}
\begin{table}
\caption{Validating the effectiveness of the diagnostic omission writing system in a tertiary hospital in China}
\label{Validating the effectiveness of the diagnostic omission writing system in a tertiary hospital in China}
\centering
\begin{tabular}{|p{70pt}|p{70pt}|p{70pt}|}
\hline
 number of EMRs & number of case with missed diagnoses & number of CC/MCC\\ 
\hline

9385 &  5116&2532\\
\hline
number of modified DRG groups & cost avoided (:RMB) & \% of cost loss avoided\\
\hline
366&1,241,030&7.49\% \\
\hline

\end{tabular}
\label{tab1}
\end{table}
After cooperating with a tertiary hospital that pays based on DRG grouping, we obtained a total of about 10,000 copies of its data for a certain month in 2021, including electronic medical record data and medical insurance settlement data. The medical insurance settlement data is the medical record grouping result and the corresponding reimbursement amount returned by the medical insurance bureau to the hospital after the hospital submits the medical record to the medical insurance bureau. We take this medical record data as input, run the diagnostic miss-write detection system and get the results. The diseases that were missed were coded by ICD, and it was judged whether the disease was CC/MCC (complication/serious complications), and compared with the DRG group returned by the medical insurance bureau. If the CC/MCC level of the missing writing disease is higher than the CC/MCC level of the original DRG grouping, a modified DRG grouping is obtained based on the disclosed CHS-DRG grouping scheme. By doing this we obtained DRG groupings adjusted for diagnosing missing write systems. The CC/MCC table is published together with the DRG group.

Since some weights change with months and we cannot obtain them, we use the average cost of each DRG group to estimate the medical insurance reimbursement costs. Through the average cost of the DRG group published by the local medical insurance bureau, we subtract the adjusted reimbursement fee for each medical record from the reimbursement fee provided by the medical insurance bureau to obtain the difference for each medical record. Summing up the differences for each medical record obtains the cost savings for the one-month medical record due to the missed diagnosis system. Since the diseases that were missed from the discharge diagnosis have been confirmed in the medical records, it is completely reasonable and legal for the recall of diseases to lead to the adjustment of CC/MCC and the increase of reimbursement expenses.

The experimental data are shown in Table \ref{Validating the effectiveness of the diagnostic omission writing system in a tertiary hospital in China}. Of course, due to the problem of accuracy, the actual adjustment of the cost is slightly lower(92.5\% accuracy). It can be seen from the experimental results that our diagnostic missed writing system will lead to a large cost loss (about 7\%) for the hospital recall due to missed diagnosis. This also helps to make the DRG grouping accurate and cost-effective for correct test cases, which in turn is beneficial to the overall DRG grouping system. This also verifies that the realization of the system for diagnosing missing writing has great economic and social benefits. Although our experiments were conducted on Chinese electronic medical records, our proposed framework is also adapted to English electronic medical records. Because of data, tools, and data annotations, we did not conduct the English electronic medical record experiment.

\section{Conclusion}
To the best of our knowledge, it is the first time that a write-missing diagnosis detection framework can realize omission detection for clearly diagnosed diseases in medical records. Although additional diagnosis is widely regarded as a key element that easily leads to grouping errors under the DRG system, it is also a key issue in the quality inspection of medical records. However, there has been no research on automatic detection using natural language processing technology in academia. In fact, at present, the missed writing of additional diagnoses mainly relies on coders or clinicians to read the whole medical record sentence by sentence, which is a very expensive method. Based on natural language methods, this problem can be effectively solved but has not been fully studied by the academic community. For the first time, we propose an effective solution framework, verify the feasibility of solving this problem, and demonstrate the social value and economic value that can be generated by solving this problem. At the same time, this is also a typical case of artificial intelligence methods effectively solving practical social problems.

Existing frameworks mainly focus on the detection of missed diagnoses of clearly appearing diseases, but it is difficult to achieve the ability to infer missed diagnoses based on the whole medical record as intelligently as a clinician. Obtaining this ability requires the realization of interpretable and traceable auxiliary diagnosis ability, which is the bottleneck of medical natural language processing at present. We will continue to research in this direction to come up with a more intelligent and comprehensive method for detecting missed write detection.

\bibliographystyle{unsrt}
\bibliography{references}

\end{CJK*}
\end{document}